\documentclass[letterpaper, 10 pt, conference]{ieeeconf}  

\IEEEoverridecommandlockouts                              

\usepackage{amsmath,amsfonts,amssymb,dsfont}
\usepackage{booktabs}
\usepackage{graphicx}
\usepackage{cite}
\usepackage{multirow}
\usepackage{float}
\usepackage{color}
\usepackage{hyperref}
\usepackage{array}
\usepackage{pifont}
\hypersetup{
    colorlinks=false,
    pdfborder={0 0 0},
}
\usepackage[table]{xcolor}

\newcolumntype{P}[1]{>{\centering\arraybackslash}p{#1}}
\definecolor{BlueHighlight}{HTML}{E5F1FF}
\newcommand{\xmark}{\ding{55}}

\title{\LARGE \bf
MDHA: Multi-Scale Deformable Transformer with Hybrid Anchors for Multi-View 3D Object Detection
}

\author{Michelle Adeline$^{1}$, Junn Yong Loo$^{1,2}$, Vishnu Monn Baskaran$^{1}$
\thanks{$^{1}$The authors are with the School of Information Technology, Monash University Malaysia 
(e-mail: \url{made0008@student.monash.edu}, \url{[loo.junnyong,vishnu.monn]@monash.edu)}.}
\thanks{$^{2}$Corresponding author}%
}
\begin{document}

\maketitle
\thispagestyle{empty}
\pagestyle{empty}

\begin{abstract}
Multi-view 3D object detection is a crucial component of autonomous driving systems. Contemporary query-based methods primarily depend either on dataset-specific initialization of 3D anchors, introducing bias, or utilize dense attention mechanisms, which are computationally inefficient and unscalable. To overcome these issues, we present MDHA, a novel sparse query-based framework, which constructs adaptive 3D output proposals using hybrid anchors from multi-view, multi-scale image input. Fixed 2D anchors are combined with depth predictions to form 2.5D anchors, which are projected to obtain 3D proposals. To ensure high efficiency, our proposed Anchor Encoder performs sparse refinement and selects the top-$k$ anchors and features. Moreover, while existing multi-view attention mechanisms rely on projecting reference points to multiple images, our novel Circular Deformable Attention mechanism only projects to a single image but allows reference points to seamlessly attend to adjacent images, improving efficiency without compromising on performance. On the nuScenes val set, it achieves 46.4\% mAP and 55.0\% NDS with a ResNet101 backbone. MDHA significantly outperforms the baseline where anchor proposals are modelled as learnable embeddings. Code is available at \url{https://github.com/NaomiEX/MDHA}.

\end{abstract}

\section{INTRODUCTION}
Multi-view 3D object detection plays a pivotal role in mapping and understanding a vehicle's surroundings for reliable autonomous driving. Among existing methods, camera-only approaches have gained immense traction  as of late due to the accessibility and low deployment cost of cameras as opposed to conventional LiDAR sensors. Camera-only methods can be split into: Bird's-Eye-View (BEV) methods \cite{lss, bevdet, bevdet4d, bevdepth , bevformer, bevformerv2} where multi-view features are fused into a unified BEV representation, and query-based methods \cite{detr3d,petr,liu2023petrv2,focalpetr, StreamPETR, sparse4d, sparse4dv2, sparse4dv3, simmod, MV2D, far3d} where 3D objects are directly modelled as queries and progressively refined based on image features.

The construction of BEV maps from input features involves a non-trivial view transformation, rendering it computationally expensive. Moreover, BEV representations suffer from having a fixed perception range, constraining their adaptability to diverse driving scenarios. For instance, in scarcely populated rural areas with minimal visual elements of interest, constructing these dense and rich BEV maps is a waste of computational resources. In contrast, query-based methods bypass the computationally intensive BEV construction and have recently achieved comparable performance to BEV-based methods with straightforward sparsification.

\begin{figure}[htb]
{\centering\includegraphics[width=\linewidth]{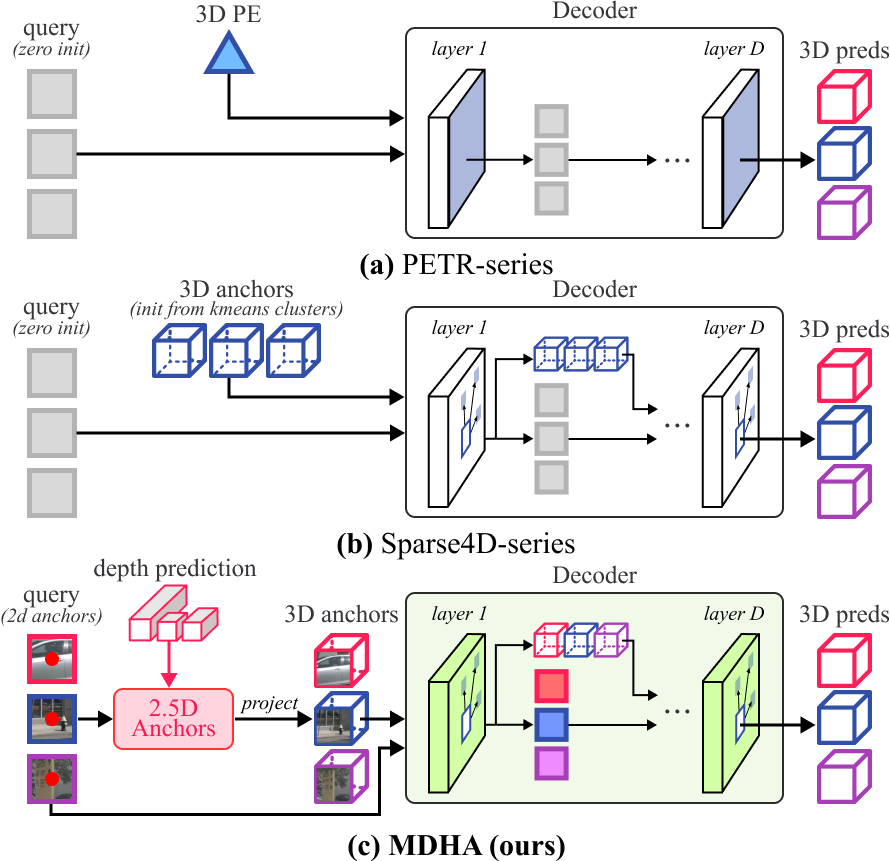}\par}
\caption{\textbf{Comparison between PETR and Sparse4D models with our proposed architecture.} (a) PETR models encode 3D information via positional embeddings and refine queries using dense attention within their decoder. (b) Sparse4D models initialize 3D anchors from k-means clustering on nuScenes and iteratively refine both queries and anchors using sparse attention within the decoder. (c) MDHA employs image tokens as queries; the $(x,y)$ center coordinates of each token, paired with depth predictions, form 2.5D anchors, which are projected into 3D anchors. This eliminates the need for good anchor initialization. Anchors and queries undergo iterative refinement using multi-view-spanning sparse attention within the decoder.}
\label{fig:model_comp}
\end{figure}
Two predominant series of models have emerged within query-based approaches: the PETR-series \cite{petr, liu2023petrv2, focalpetr, StreamPETR}, which generates 3D position-aware features from 2D image features via positional embeddings, and the Sparse4D-series \cite{sparse4d, sparse4dv2, sparse4dv3}, which refines 3D anchors using sparse feature sampling of spatio-temporal input. Sparse methods are attractive since they enable easy tuning to achieve the desired balance between performance and efficiency for real-world application. However, since PETR models employ a dense cross-attention mechanism, they cannot be considered as sparse methods. Moreover, they do not utilize multi-scale input features, limiting their scalability to detect objects of varying sizes. Although these are rectified by the Sparse4D models, they instead rely on anchors initialized from k-means clustering on the nuScenes \cite{nuscenes} train set, thus potentially compromising their ability to generalize to real-world driving scenarios.

To address these issues, we propose a novel framework for camera-only query-based 3D object detection centered on hybrid anchors. Motivated by the effectiveness of 2D priors in 3D object detection \cite{bevformerv2, simmod, MV2D, far3d}, we propose the formation of 2.5D anchors by pairing each token's 2D center coordinates with corresponding depth predictions. These anchors can be projected, with known camera transformation matrices, to generate reasonable 3D output proposals. Our usage of multi-scale, multi-view input leads to a large number of tokens, and thus, of 3D proposals, posing significant computational burden for refinement. To alleviate this, our proposed Anchor Encoder serves the dual-purpose of refining and selecting top-$k$ image features and 3D proposals for subsequent iterative refinement within the spatio-temporal MDHA Decoder. On top of this, to improve efficiency, we propose a novel Circular Deformable Attention (CDA) mechanism, which treats multi-view input as a contiguous 360$^{\circ}$ panoramic image. This allows reference points to seamlessly attend to locations in adjacent images. Thus, our proposed method eliminates the reliance on good 3D anchor initialization, leverages multi-scale input features for improved detection at varying scales, and improves efficiency through our novel sparse attention mechanism, which offers greater flexibility than existing multi-view attention mechanisms, where attention is confined to the image in which the reference point is projected. Figure \ref{fig:model_comp} illustrates the comparison between our method and the PETR and Sparse4D models.

To summarize, our main contributions are as follows:
\begin{itemize}
    \item We propose a novel framework for sparse query-based 3D object detection from multi-view cameras, MDHA, which constructs adaptive and diverse 3D output proposals from 2D$\rightarrow$2.5D$\rightarrow$3D anchors. Top-$k$ proposals are sparsely selected in the Anchor Encoder and refined within the MDHA decoder, thereby reducing reliance on 3D anchor initialization.
    \item An elegant multi-view-spanning sparse attention mechanism, Circular Deformable Attention, which improves efficiency without compromising performance.
    \item On the nuScenes val set, MDHA significantly outperforms the learned anchors baseline, where proposals are implemented as learnable embeddings, and surpasses most state-of-the-art query-based methods.
\end{itemize}

\section{RELATED WORK}
\begin{figure*}[ht]
    \centering
    \includegraphics[width=1.0\linewidth]{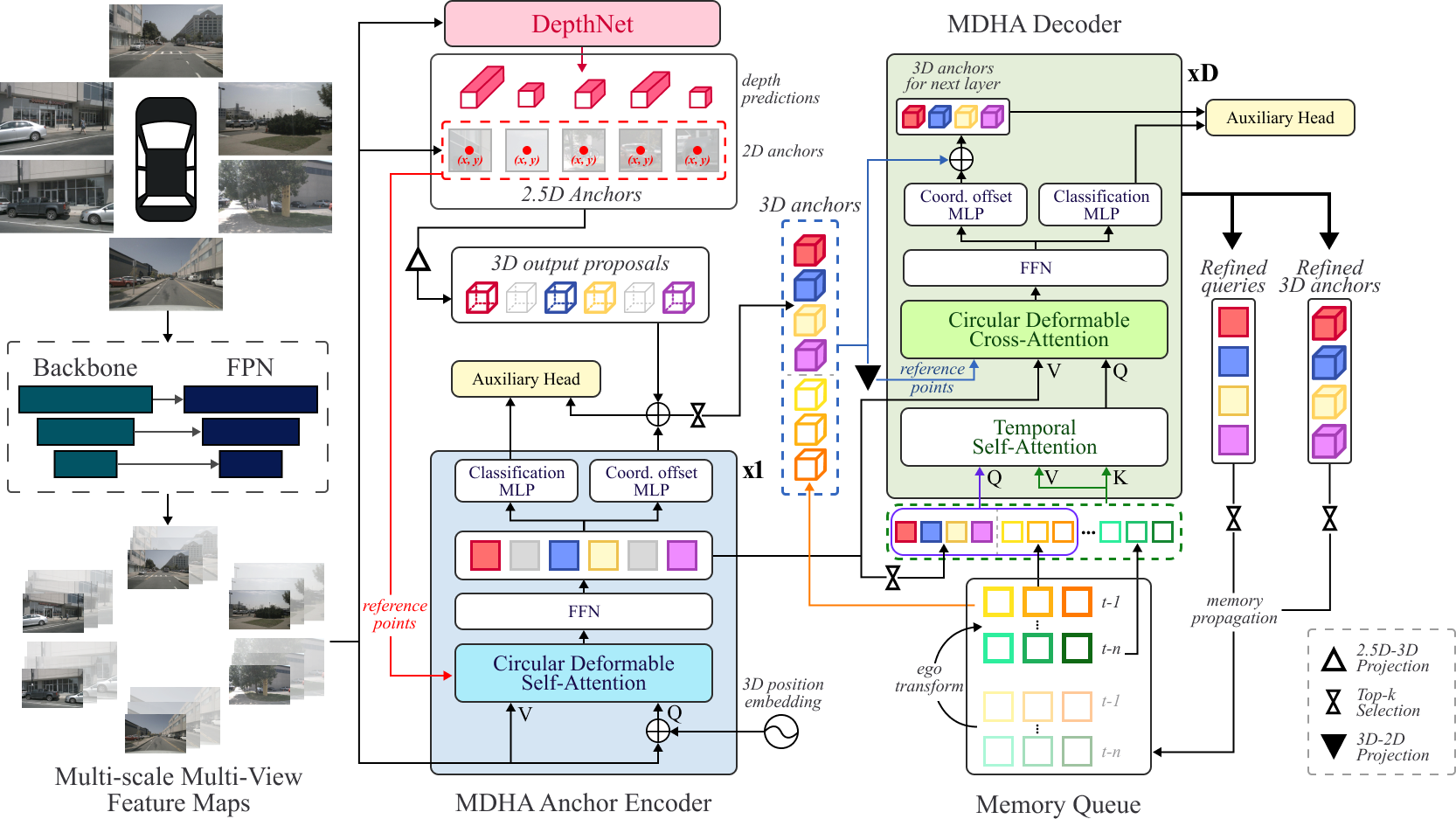}
    \caption{\textbf{Our proposed MDHA Architecture}. Multi-view images are fed into the backbone and FPN neck to extract multi-scale, multi-view image features. These feature tokens serve as input for the DepthNet, which pairs their 2D center coordinates with predicted depth, forming 2.5D anchors, which are projected to obtain 3D output proposals. The tokens and proposals undergo refinement in the 1-layer MDHA Anchor Encoder. It also selects the top-$k$ queries and proposals for further refinement in the MDHA Decoder, which additionally considers temporal information via the memory queue.}
    \label{fig:model_diag}
\end{figure*}

\subsection{Camera-Only Multi-View 3D Object Detection}

BEV-based methods perform 3D object detection by leveraging a Bird's-Eye-View feature representation acquired by transforming image features. In most works \cite{bevdepth,lss,bevdet, bevdet4d}, this transformation follows the Lift-Splat paradigm \cite{lss}, which involves ``lifting'' image features into 3D space using depth predictions, and ``splatting'' them onto the BEV plane by fusing features that fall into the pre-defined grids. BEVDet \cite{bevdet} constructs a BEV map through this view transformation, refines it with a BEV-Encoder and generates predictions with a detection head. BEVDet4D \cite{bevdet4d} extends this framework by incorporating temporal features which are projected onto the current frame, while BEVDepth \cite{bevdepth} introduces explicit depth supervision with a camera-aware depth estimation module. Meanwhile, BEVFormer \cite{bevformer} models BEV pillars as dense queries and utilizes deformable attention to aggregate spatio-temporal information for BEV refinement.



In contrast, query-based methods circumvent the complex construction of BEV maps by modelling objects implicitly as queries. DETR3D \cite{detr3d} spearheaded this class of methods by learning a 3D-to-2D projection of predicted 2D object centers, yielding reference points which, in conjunction with sampled image features, are employed for query refinement in the decoder. Despite being a representative sparse query-based method, it suffers from poor performance as it neglects temporal information. Sparse4D \cite{sparse4d} rectifies this by projecting 4D keypoints onto multi-view frames across multiple timestamps. Sparse4Dv2 \cite{sparse4dv2} improves both its performance and efficiency by adopting a recurrent temporal feature fusion module. PETR \cite{petr} diverges from DETR3D by encoding 3D spatial information into input features via positional embedding, eliminating the need for 3D-to-2D projection. PETRv2 \cite{liu2023petrv2} extends this framework to other 3D perception tasks, namely BEV segmentation and lane detection, while Focal-PETR \cite{focalpetr} adds a focal sampling module which selects discriminative foreground features and converts them to 3D-aware features via spatial alignment. Finally, StreamPETR \cite{StreamPETR} demonstrates impressive performance by adopting an object-centric temporal fusion method which propagates top-$k$ queries and reference points from prior frames into a small memory queue for improved temporal modelling. Inspired by the aforementioned models, our query-based method adopts object-centric recurrent temporal fusion and employs 3D-to-2D projections in our attention mechanism. However, our method uniquely builds up adaptive 3D output proposals from image features rather than using learnable embeddings.



\subsection{Depth and 2D Auxiliary Tasks for 3D Object Detection}
In an emerging trend, more frameworks are integrating depth or 2D modules for auxiliary supervision \cite{focalpetr, StreamPETR, sparse4dv2, sparse4dv3} or for the final 3D prediction \cite{bevformerv2, simmod, far3d, MV2D}.
Sparse4Dv2 \cite{sparse4dv2} and Sparse4Dv3 \cite{sparse4dv3} both implement auxiliary dense depth supervision to improve training stability while Focal-PETR \cite{focalpetr} and StreamPETR \cite{StreamPETR} both utilize auxiliary 2D supervision for the same purpose.

In contrast, certain methods have integrated the outputs of these auxiliary modules into the final 3D prediction. For instance, BEVFormer v2 \cite{bevformerv2} proposes a two-stage detector, featuring a perspective head that suggests proposals used within the BEV decoder. SimMOD \cite{simmod} generates proposals for each token via four convolutional branches, predicting the object class, centerness, offset, and depth, where the first two are used for proposal selection and the latter two predict the 2.5D center. MV2D \cite{MV2D} employs a 2D detector to suggest regions of interest, from which aligned features and queries are extracted for its decoder. Far3D \cite{far3d} focuses on long-range object detection by constructing 3D adaptive queries from 2D bounding box and depth predictions.
Our approach differs from existing works by streamlining the auxiliary module, requiring only depth predictions. By pairing depth values with the 2D coordinates of each feature token, we bypass the need to predict 2D anchors, reducing the learning burden of the model and improving efficiency. Furthermore, proposal selection, feature aggregation, and sparse refinement are all executed by a shallow transformer encoder.

\section{METHOD}
\label{sec:method}
\subsection{Overview}
Figure \ref{fig:model_diag} shows the overall architecture of MDHA. Given $N$-view RGB images, the backbone and Feature Pyramid Network (FPN) neck extracts multi-scale feature maps, $\{F_l\}^{L}_{l=1}$, where $F_l\in \mathbb{R}^{N\times C\times H_l\times W_l}$, $C$ denotes the feature dimension, and $(H_l, W_l)$ refers to the height and width of the feature map at level $l$. For each feature token (feature map cell), the DepthNet constructs 2.5D anchors, which are projected to obtain 3D output proposals (Section \ref{ssec:depthnet}). The single-layer MDHA Anchor Encoder refines and selects the top-$k$ features and proposals to pass onto the decoder (Section \ref{ssec:encoder}). The $D$-layer MDHA Decoder conducts iterative anchor refinement using spatio-temporal information (Section \ref{ssec:decoder}). Within the model, CDA is employed as an efficient multi-view-spanning sparse attention mechanism (Section \ref{ssec:cda}). The model is trained end-to-end with detection and classification (final and auxiliary) losses, with explicit depth supervision (Section \ref{ssec:loss}).
\subsection{DepthNet}
\label{ssec:depthnet}
Constructing reasonable proposals for 3D object detection is not an easy feat. While a naive approach involves parameterizing initial anchors, making them learnable, the large search space makes this a non-trivial task, and sub-optimal anchors run the risk of destabilizing training and providing inadequate coverage of the perception range (Section \ref{ssec: hybrid_vs_learnable_anchors}). Although Sparse4D avoids this issue by initializing 900 anchors from k-means clustering on the nuScenes train set, this introduces bias towards the data distribution present in nuScenes. Thus, with the aim of generating robust proposals for generalized driving scenarios, we opt to use the 2D center coordinates of each feature token to construct a set of 2.5D anchors. For the $i$-th token, we define the 2.5D anchor as:
\begin{equation}
A^{2.5D}_i=[x^{2D}_i, y^{2D}_i, z_i]^T 
\end{equation}
where $(x^{2D}_i, y^{2D}_i)=(\frac{x_i + 0.5}{W_l} \cdot W_{inp}, \frac{y_i + 0.5}{H_l} \cdot H_{inp})$ are the 2D coordinates of the token's center, ${z_i}$ is the predicted depth of the object center, and $(W_{inp}, H_{inp})$ is the input image resolution. By fixing the $(x,y)$ coordinates, the model only focuses on learning the depth value, reducing the search space from $\mathbb{R}^3$ to $\mathbb{R}^1$. Furthermore, since each feature token covers a separate part of the image, the resulting 3D anchors are naturally well-distributed around the ego vehicle.

These 2.5D anchors can then be projected into 3D output proposals using $E_i$ and $I_i$, the camera extrinsic and intrinsic matrices, respectively, for the view in which the token belongs to:
\begin{equation}
A^{3D}_i = E_i^{-1} I_i^{-1} [x^{2D}_i \ast z_i, y^{2D}_i \ast z_i, z_i, 1]^T
\end{equation}

\begin{figure}[htb]
{\centering\includegraphics[width=\linewidth]{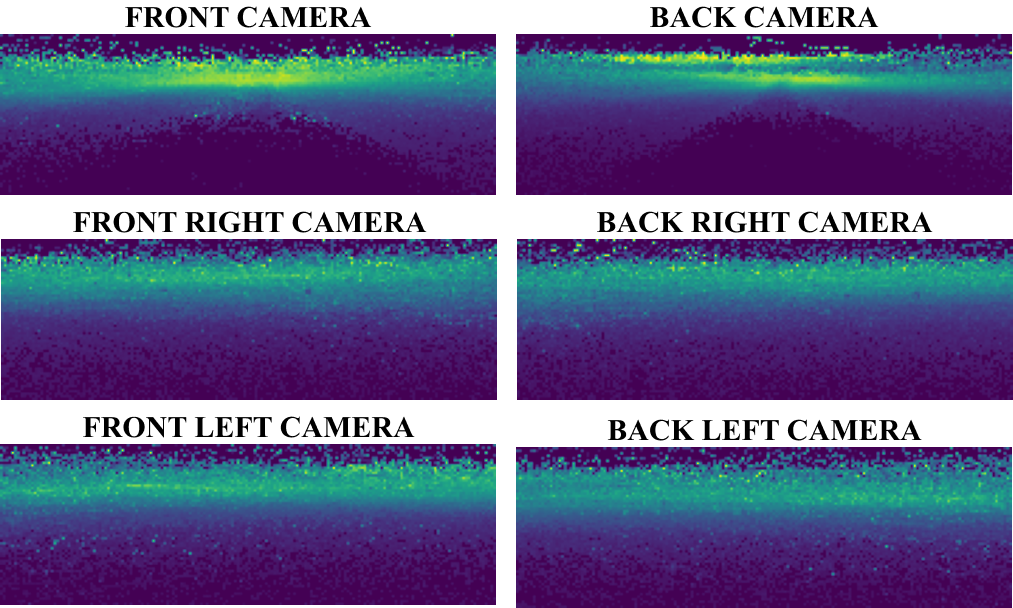}\par}
\caption{\textbf{Depth distribution of 3D object centers projected onto all 6 cameras in the nuScenes train set.} The brighter the colour, the farther away the object. The semicircular voids at the bottom of the front and back cameras represent the front and rear part of the vehicle which juts out and where no objects can be located.}
\label{fig:depth_dist}
\end{figure}

The DepthNet assumes that an object is centered within a token and predicts its depth relative to the ego vehicle. Below, we examine two approaches in obtaining depth suggestions:

\textbf{Fixed Depth.} Figure \ref{fig:depth_dist} shows a clear pattern in the depth distribution  within each camera: depth increases as we travel up the image. Motivated by this observation, we sample $z_i$ from this distribution, eliminating the need for the model to predict the depth entirely. We argue that this is generalizable since, in most driving scenarios, all objects are situated on a level plane. Hence, for an object center to appear higher up on the image, it must be further away from the ego vehicle.

\textbf{Learnable Depth.} We also explore a more adaptive approach for obtaining depth maps, $\{z_l\}_{l=1}^{L}$, where $z_l\in\mathbb{R}^{N\times C\times H_l\times W_l}$ is predicted as follows:
\begin{equation}
z_l = \sigma(\boldsymbol{\Psi}(F_l)) \ast (D^{max}-D^{min}) + D^{min}
\end{equation}
where $\boldsymbol{\Psi}$ is a single layer convolutional head and $D^{max}$, $D^{min}$ are the maximum and minimum depth values.



\subsection{MDHA Anchor Encoder}
\label{ssec:encoder}
Within the encoder, each image token acts as a query and undergoes refinement via self-attention using our sparse CDA mechanism, where $(x^{2D}_i, y^{2D}_i)$ serve as 2D reference points for each query. The addition of 3D position embeddings from PETR was found to stabilize training by providing 3D positional context for the 2D image tokens. We also employ $(x^{2D}_i, y^{2D}_i, z_i)$ sinusoidal position embeddings to encode the token's learned 2.5D anchor. For each token, two Multilayer Perceptron (MLP) heads predict the object classification score and offsets to refine the token's corresponding 3D proposal. The resulting proposals with the top-$k$ classification scores, $A^{3D}_{enc}$, are selected for further refinement within the decoder. We follow DINO \cite{dino} in initializing the decoder queries as the top-$k$ queries from the encoder, $Q_{enc}$.



\subsection{MDHA Decoder}
\label{ssec:decoder}
The decoder utilizes temporal information by leveraging a short memory queue consisting of sparse features and anchors from the last $m$ frames. To account for ego vehicle movement between frames, historic 3D anchors, $A^{3D}_{t-\tau}=(x^{3D}, y^{3D}, z)_{t-\tau}$, are aligned to the current frame as follows:
\begin{equation}
A^{3D}_{t-\tau\rightarrow t}=\text{EGO}^{-1}_{t}\cdot\text{EGO}_{t-\tau} [A_{t-\tau}^{3D}, 1]^T
\end{equation}
where $\text{EGO}_t$ is the lidar-to-global transformation matrix at time $t$.

Relevant historic features are then selected via the Temporal Self-Attention module which is implemented as vanilla multi-head attention \cite{transformer} where the features from the memory queue serve as the $key$ and $value$, while the $query$ consists of $(Q_{enc}, Q_{t-1})$, with $Q_{t-1}$ being the query propagated from the previous frame.  Due to the fixed-length memory queue, this operation's computational complexity does not scale with increasing image resolution and since memory size $<<$ feature map size, it is dominated by the encoder's complexity which \textit{does} scale with feature map size. Thus, sparse attention is unnecessary for this operation.

In the Circular Deformable Cross-Attention module, these selected queries efficiently attend to refined image features from the encoder via our CDA mechanism. Its reference points for decoder layer $d$, are obtained by projecting 3D anchors, $A^{d-1}$, from the previous layer, into 2D. For the first layer, these anchors are defined as $A^0=[A^{3D}_{enc}, A^{3D}_{t-1 \rightarrow t}]$, and the 3D-to-2D projection for view $n$ is given by: 
\begin{equation}\label{eq:proj_3d_2d}
    \text{Proj}(A^{3D}, n) = I_n \cdot E_n [A^{3D}, 1]^T
\end{equation}


Since a 3D point can be projected to multiple camera views, we either contend with a one-to-many relationship between queries and reference points or choose a single projected point via a heuristic. For this work, we opt for the latter as it is more efficient, and with our novel CDA mechanism, performance is not compromised. Our chosen heuristic involves selecting the point that is within image boundaries and is closest to the center of the view it is projected. Thus, our reference points are given by:
\begin{equation}
(r^d_x, r^d_y)=\underset{(r_x, r_y)\in \mathcal{R}}{\arg\min} \Vert[r_x, r_y]^T - [W_{inp}/2, H_{inp}/2]^T\Vert_2
\end{equation}
where $\mathcal{R}=\{\text{Proj}(A^{d-1}, n)\}^N_{n=1}\setminus\mathcal{R}_{\oslash}$ is the set of projected 2D points, excluding points outside image boundaries, $\mathcal{R}_{\oslash}$.

For each query, two MLP heads predict classification scores and offsets, $\Delta A^d$, which are used to obtain the refined anchors $A^d = A^{d-1} + \Delta A^d$ for each decoder layer. Anchors and queries with the top-$q$ scores are propagated into the memory queue. For improved efficiency, intermediate classification predictions are omitted during inference time.


\subsection{Circular Deformable Attention (CDA)}
\label{ssec:cda}
Self-attention in the encoder and cross-attention in the decoder utilize multi-scale, multi-view image features as attention targets. Due to the large number of feature tokens, executing these operations using vanilla multi-head attention is computationally expensive. Instead, we employ a modification of the deformable attention mechanism \cite{deformabledetr}. A straightforward implementation of deformable attention in a multi-view setting is to treat each of the $N$ views as separate images within the batch, then project each 3D anchor into one or more 2D reference points spread across multiple views as is done in most existing works \cite{detr3d,sparse4d,sparse4dv2,sparse4dv3}. However, this approach limits reference points to only attend to locations within their projected image. We overcome this limitation by concatenating the $N$ views into a single contiguous 360${^{\circ}}$ image. Thus, our feature maps are concatenated horizontally as $\mathcal{M}=\{M_l\}^L_{l=1}$, where $M_l=[F_{l1}, F_{l2}, \dots, F_{lN}]$ and $F_{ln}$ represents the input features at level $l$ for view $n$.

Given 2D reference points, $r_x \in [0,W_{inp}], r_y \in [0,H_{inp}]$ local to view $n$, we can obtain normalized global 2D reference points, $\hat{r}^{cda}=(\hat{r}^{cda}_x, \hat{r}^{cda}_y)\in [0,1]^2$, for $\mathcal{M}$ as follows:
\begin{equation}
\hat{r}^{cda}_x = \frac{r_x + (n-1)W_{inp}}{N\times W_{inp}},\ \hat{r}^{cda}_y = \frac{r_y}{H_{inp}}
\end{equation}
Letting $q_i$ be the $i$-th query, CDA is then formulated as:
\begin{equation}
\text{CDA}(q_i, \hat{r}^{cda}_i, \mathcal{M}) = \sum^{N_{h}}_{h=1} W_{h} \sum^L_{l=1} \sum^S_{s=1} A_{hlis} \cdot W'_h x_l(p_{hlis})
\end{equation}
where $h$ indexes the attention head with a total of $N_h$ heads, and $s$ indexes the sampling location with a total of $S$ locations. $W_h\in \mathbb{R}^{C\times(C/N_h)}$, $W'_h\in \mathbb{R}^{(C/N_h) \times C}$ are learnable weights, and $A_{hlis}\in[0,1]$ is the predicted attention weight. Here, $x_l(p_{hlis})$ refers to the input feature sampled via bilinear interpolation from sampling location $p_{hlis}=\phi(\hat{r}^{cda}_i + \Delta p_{hlis})$, where $\phi$ scales the value to the feature map's resolution, and $\Delta p_{hlis}$ is the sampling offset obtained as follows:
\begin{equation}
\Delta p_{hlis}=\boldsymbol{\Phi}(q_i) \,/\, (W^{\mathcal{M}}_l, H^{\mathcal{M}}_l)
\end{equation}
with $\boldsymbol{\Phi}$ denoting the linear projection. Since $(W^{\mathcal{M}}_l, H^{\mathcal{M}}_l)=(N \times W_l, H_l)$, $\Delta p_{hlis}$ is not bound by the dimensions of the view it is projected onto, inherently allowing the model to learn sampling locations beyond image boundaries.
Considering that the $\boldsymbol{\Phi}(q_i)$ is unbounded, the offset might exceed the size of the feature map at that level. Therefore, we wrap sampling points around as if the input were circular:
\begin{equation}
\mathring{p}_{hlis}=p_{hlis}\; \text{mod}\; 1.0
\end{equation}
In effect, this treats the first and last view as if they are adjacent. CDA works best if visual continuity is maintained between neighbouring images. Thus, we reorder inputs to follow a circular camera order, i.e., Front\;$\rightarrow$\;Front-Right\;$\rightarrow$\;Back-Right\;$\rightarrow$\;Back\;$\rightarrow$\;Back-Left\;$\rightarrow$\;Front-Left. This arrangement places related features next to one another. No other camera calibrations were done.



\begin{figure}[!h]
{\centering\includegraphics[width=\linewidth]{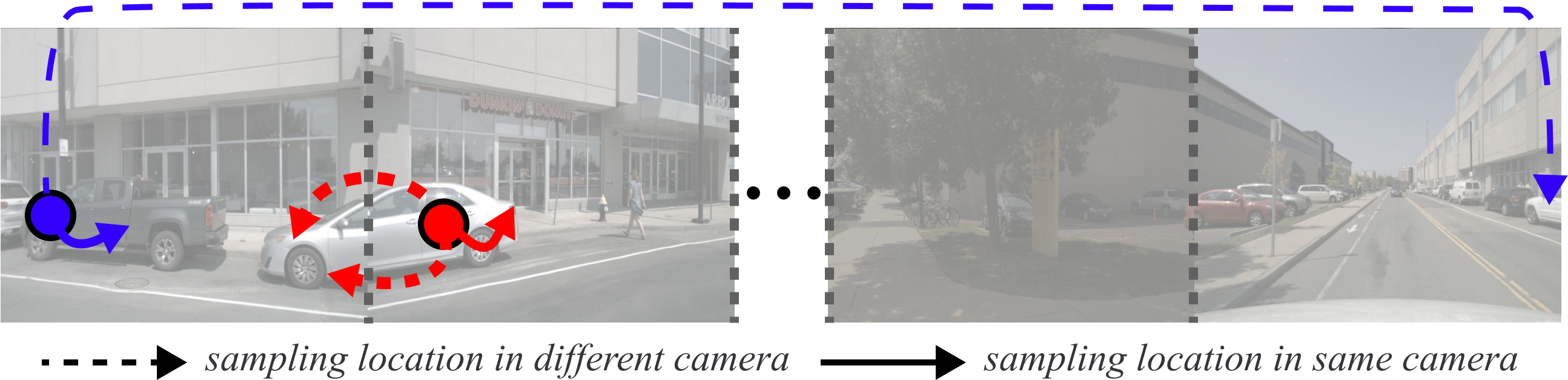}\par}
\caption{\textbf{CDA on horizontally concatenated input}. The circles denote reference points and the arrows represent sampling locations.}
\label{fig:cda}
\end{figure}

Since autonomous vehicles require a 360$^{\circ}$ view for safe navigation, CDA is highly viable for real-world application. Furthermore, given that CDA only involves reshaping input features and inexpensive pre-processing of reference points, it adds negligible overhead on top of vanilla deformable attention. Figure \ref{fig:cda} illustrates the CDA mechanism.




\subsection{Training Loss}
\label{ssec:loss}
We define a 3D detection as follows:
\[ \begin{bmatrix} x^{3D} ,\; y^{3D} ,\; z ,\; w ,\; l ,\; h ,\; \theta ,\; v^x ,\; v^y \end{bmatrix} \]
consisting of the object's 3D center, width, length, height, yaw, and its $x$ and $y$ velocity. Let $\{\tilde{b}_j\}^{N_{gt}}_{j=1}$ and $\{\tilde{z}_j\}^{N_{gt}}_{j=1}$ be the set of ground-truth 3D detections and depths, while $\{\hat{b}_i\}^{N_q}_{i=1}$ and $\{\hat{z}_k\}^{N_{tok}}_{k=1}$ represent the set of predicted 3D detections and depths, where $N_{tok} = \sum^L_{l=1} N\times H_l\times W_l$.
Under this setting, MDHA is trained end-to-end to minimize:
\begin{equation}\label{eq:MDHA_loss}
\mathcal{L}=\lambda_1 \mathcal{L}_{cls} + \lambda_2 \mathcal{L}_{det} + \lambda_3 \mathcal{L}_{depth}
\end{equation}
where the classification loss, $\mathcal{L}_{cls}$, is implemented using focal loss \cite{focalloss}, and the detection loss is defined as 
$\mathcal{L}_{det}=\frac{1}{N_{gt}}\sum^{N_q}_{i=1}\mathds{1}_{\{b^{\text{targ}}_i\neq\varnothing\}}|b^{\text{targ}}_i - \hat{b}_i|$, with the chosen ground-truth target $b^{\text{targ}}_i$ of prediction $\hat{b}_i$ obtained via bipartite matching \cite{detr}. Auxiliary classification and detection losses are used in the encoder and intermediate decoder layers. To calculate the depth loss $\mathcal{L}_{depth}$, we project $\tilde{b}_j$ to all $N$ views using (\ref{eq:proj_3d_2d}) to obtain $\{(\tilde{x}^{2D}_{jn}, \tilde{y}^{2D}_{jn})\}^{N^{proj}_j}_{n=1}$, where $N^{proj}_j=N-N_{\oslash}$ and $N_{\oslash}$ refers to the number of projected 2D points outside image boundaries. The target of depth prediction $\hat{z}_k$ is denoted as $\tilde{z}_{\hat{m}}$, with index $\hat{m}$ obtained via:
\begin{equation}
\hat{m}=\underset{j\in [1, N_{gt}]}{\arg\min} \; D_{kj}
\end{equation}
where $D_{kj}= \underset{n}{\min}\Vert[\hat{x}^{2D}_k, \hat{y}^{2D}_k]^T -  [\tilde{x}^{2D}_{jn}, \tilde{y}^{2D}_{jn}]^T\Vert_1$. Finally, we define the depth loss in (\ref{eq:MDHA_loss}) as 
\begin{equation}
\mathcal{L}_{depth}=\frac{1}{\sum^{N_{gt}}_{j=1} N^{proj}_j \times L}\sum^{N_{tok}}_{k=1}W_k\lvert\tilde{z}_{\hat{m}} - \hat{z}_k\rvert
\end{equation}
Let $W^{decay}_k=e^{-D_{k\hat{m}}/\varepsilon}$ represent the exponentially decayed weight of prediction $k$, which decreases the further away the token's 2D coordinates are from the target ground-truth object. The final weight of prediction $k$ is obtained by imposing a strict distance cutoff:
\begin{equation}
    W_k=
    \begin{cases}
      W^{decay}_k, & \text{if}\ W^{decay}_k>\rho \\
      0, & \text{otherwise}
    \end{cases}
\end{equation}
to ensure that only predictions in close proximity to ground truth objects propagate gradients, striking a balance between incredibly sparse one-to-one matching and exhaustive gradient propagation, for stable training.

\section{EXPERIMENTS}
\begin{table*}[h]
\begin{center}
\caption{Performance comparison on the nuScenes val set. $^{\ast}$ is trained with CBGS. $^{\dagger}$ uses pre-trained weights from FCOS3D \cite{fcos3d}. $^{\ddagger}$ benefited from pre-training on nuImages. $^{\mathsection}$ uses 900 anchors initialized from kmeans clustering of the nuScenes train set.\label{tab:nusc_val_res}}
\begin{tabular}{
p{0.13\textwidth}| P{0.11\textwidth}| P{0.08\textwidth}| P{0.06\textwidth} P{0.04\textwidth}| 
P{0.0418\textwidth} P{0.0418\textwidth} P{0.0418\textwidth} P{0.0418\textwidth}  P{0.0418\textwidth} P{0.0418\textwidth} P{0.0418\textwidth}} 
\hline
Method & Backbone & Image Size & Sparse Attention & Frames & 
mAP$\uparrow$ & NDS$\uparrow$ &
mATE$\downarrow$ & mASE$\downarrow$ & mAOE$\downarrow$ & mAVE$\downarrow$ & mAAE$\downarrow$ \\

\hline

$\text{PETR\cite{petr}}^{\ast}$ & ResNet50-DCN &$384\times 1056$ & \xmark & 1 &
0.313 & 0.381 & 0.768 & 0.278 & 0.564 & 0.923 & 0.225 \\

$\text{Focal-PETR\cite{focalpetr} }$ & ResNet50-DCN &$320\times 800$ & \xmark & 1 &
0.320 & 0.381 & 0.788 & 0.278 & 0.595 & 0.893 & 0.228 \\

$\text{SimMOD\cite{simmod} }$ & ResNet50-DCN &$900\times 1600$ & \xmark & 1 & 0.339 & 0.432 & 0.727 & - & 0.356 & - & - \\

$\text{PETRv2\cite{liu2023petrv2} }$ & ResNet50 &$256\times 704$ & \xmark & 2 &
0.349 & 0.456 & 0.700 & 0.275 & 0.580 & 0.437 & 0.187 \\

$\text{StreamPETR\cite{StreamPETR} }$ & ResNet50 & $256\times 704$ & \xmark & 8 &
0.432 & 0.540 & 0.581 & 0.272 & 0.413 & 0.295 & 0.195 \\

$\text{Sparse4Dv2\cite{sparse4dv2}}^{\mathsection}$ & ResNet50 & $256\times 704$ & $\checkmark$ & 1 &
0.439 & 0.539 & 0.598 & 0.270 & 0.475 & 0.282 & 0.179 \\

\rowcolor{BlueHighlight}
$\text{MDHA-fixed}$ & ResNet50 & $256\times 704$ & $\checkmark$ & 1 & 0.388 & 0.497 & 0.681 & 0.277 & 0.535 & 0.301 & 0.179 \\
\rowcolor{BlueHighlight}
$\text{MDHA-conv}$ & ResNet50 & $256\times 704$ & $\checkmark$ & 1 & 0.396 & 0.498 & 0.681 & 0.276 & 0.517 & 0.338 & 0.183 \\[0.25ex]
\hline


\rule{0pt}{2.5ex}$\text{DETR3D\cite{detr3d}}^{\ast\dagger}$ & ResNet101-DCN & $900\times 1600$ & $\checkmark$ & 1 & 
0.349 & 0.434 & 0.716 & 0.268 & 0.379 & 0.842 & 0.200 \\

$\text{PETR\cite{petr} }^{\ast\dagger}$ & ResNet101-DCN & $512\times 1408$ & \xmark & 1 &
0.366 & 0.441 & 0.717 & 0.267 & 0.412 & 0.834 & 0.190 \\

$\text{Focal-PETR\cite{focalpetr} }^{\dagger}$ & ResNet101-DCN & $512\times 1408$ & \xmark & 1 &
0.390 & 0.461 & 0.678 & 0.263 & 0.395 & 0.804 & 0.202 \\

$\text{SimMOD\cite{simmod} }^{\dagger}$ & ResNet101-DCN & $900\times 1600$ & \xmark & 1 &
0.366 & 0.455 & 0.698 & 0.264 & 0.340 & 0.784 & 0.197 \\

$\text{PETRv2\cite{liu2023petrv2} }^{\dagger}$ & ResNet101 & $640\times 1600$ & \xmark & 2 &
0.421 & 0.524 & 0.681 & 0.267 & 0.357 & 0.377 & 0.186 \\

$\text{Sparse4D\cite{sparse4d} }^{\dagger\mathsection}$ & ResNet101-DCN & $900\times 1600$ & \checkmark & 4 &
0.436 & 0.541 & 0.633 & 0.279 & 0.363 & 0.317 & 0.177 \\

$\text{StreamPETR\cite{StreamPETR} }^{\ddagger}$ & ResNet101 & $512\times1408$ & \xmark & 8 &
0.504 & 0.592 & 0.569 & 0.262 & 0.315 & 0.257 & 0.199 \\

$\text{Sparse4Dv2\cite{sparse4dv2} }^{\ddagger\mathsection}$ & ResNet101 & $512\times1408$ & $\checkmark$ & 1 &
0.505 & 0.594 & 0.548 & 0.268 & 0.348 & 0.239 & 0.184 \\

\rowcolor{BlueHighlight}
$\text{MDHA-fixed}^{\ddagger}$ & ResNet101 & $512\times1408$ & $\checkmark$ & 1 & 0.451 & 0.544 & 0.615 & 0.265 & 0.465 & 0.289 & 0.182 \\
\rowcolor{BlueHighlight}
$\text{MDHA-conv}^{\ddagger}$ & ResNet101 & $512\times1408$ & $\checkmark$ & 1 & 0.464 & 0.550 & 0.608 & 0.261 & 0.444 & 0.321 & 0.184 \\
\hline

\end{tabular}
\end{center}
\end{table*}

\subsection{Implementation Details}
For fair comparison, MDHA is tested with two backbones: ResNet50 \cite{resnet} pre-trained on ImageNet \cite{imagenet} and ResNet101 pre-trained on nuImages \cite{nuscenes}. We set $D^{max}=61.2$ and $D^{min}=1.0$; a total of $N_q=900$ queries are used in the decoder, with $k=644$ queries from the encoder and $q=256$ values propagated from the previous frame. The memory queue retains sparse features and anchors from the last $m=4$ frames. We use a single-layer encoder to keep training times manageable while the decoder has $D=6$ layers. Training loss weights are set to be $\lambda_1=2.0$, $\lambda_2=0.25$, $\lambda_3=0.01$ with auxiliary losses employing the same weights. For depth loss, we set $\varepsilon=\frac{10}{l}$ and $\rho=0.01$. During training, denoising is applied for auxiliary supervision within the decoder, using 10 denoising groups per ground-truth. Following Sparse4Dv2 \cite{sparse4dv2}, the model is trained for 100 epochs for Table \ref{tab:nusc_val_res} and 25 epochs for Section \ref{sec: analysis}, both using the AdamW optimizer \cite{adamw} with 0.01 weight decay. Batch size of 16 is used with initial learning rate of 4e-4, decayed following the cosine annealing schedule \cite{cosine_anneal}. Input augmentation follows PETR \cite{petr}. No CBGS \cite{cbgs} or test time augmentation was used in all experiments.

\subsection{Dataset}
We assess MDHA's performance on the large-scale autonomous driving nuScenes \cite{nuscenes} dataset using its official performance metrics. It captures driving scenes with 6 surround-view cameras as 20-second video clips at 2 frames per second (FPS), with a total of 1000 scenes split up into 700/150/150 for training/validation/testing. The dataset is fully annotated with 3D bounding boxes for 10 object classes.

\subsection{Main Results}
Table \ref{tab:nusc_val_res} compares MDHA-conv and MDHA-fixed, which uses the Learnable and Fixed Depth approaches, respectively, against state-of-the-art camera-only query-based methods. With ResNet50, MDHA-conv outperforms most existing models, except for StreamPETR and Sparse4Dv2, which achieve slightly higher mAP and NDS. However, we would like to point out that StreamPETR benefits from dense attention and is trained with a sliding window, using 8 frames \textit{and} a memory queue, for a single prediction. In contrast, our method employs efficient and scalable sparse attention with only a window of size 1 using the same memory queue. As a result, MDHA trains \textbf{1.9$\times$ faster} than StreamPETR with the same number of epochs. Additionally, Sparse4Dv2 initializes anchors from k-means clustering on the nuScenes train set, introducing bias towards the dataset, whereas our method utilizes adaptive proposals without anchor initialization. Thus, our method does not require prior knowledge of the dataset distribution. Furthermore, we observe that despite a lower input resolution, MDHA-conv achieves 5.7 percentage points (pp) higher mAP and 6.6 pp higher NDS compared to SimMOD, another proposal-based framework which uses complex 2D priors. With a ResNet101 backbone, MDHA-conv again outperforms most existing models except for StreamPETR and Sparse4Dv2, though it does reduce mASE by 0.1 pp and 0.7 pp respectively, which could be attributed to the multi-scale feature maps and our CDA mechanism enabling queries to better reason about large or multi-view-spanning vehicles. Even with lower input resolution, MDHA-conv once again outperforms SimMOD by 9.8 pp mAP and 9.5 pp NDS. For both backbones, MDHA-fixed slightly underperforms compared to MDHA-conv due to its fixed depth distribution, but offers an inference speed of 15.1 FPS on an RTX 4090, which is 0.7 higher than MDHA-conv.

\subsection{Analysis}
\label{sec: analysis}
\begin{figure*}
    \centering
    \includegraphics[width=\linewidth]{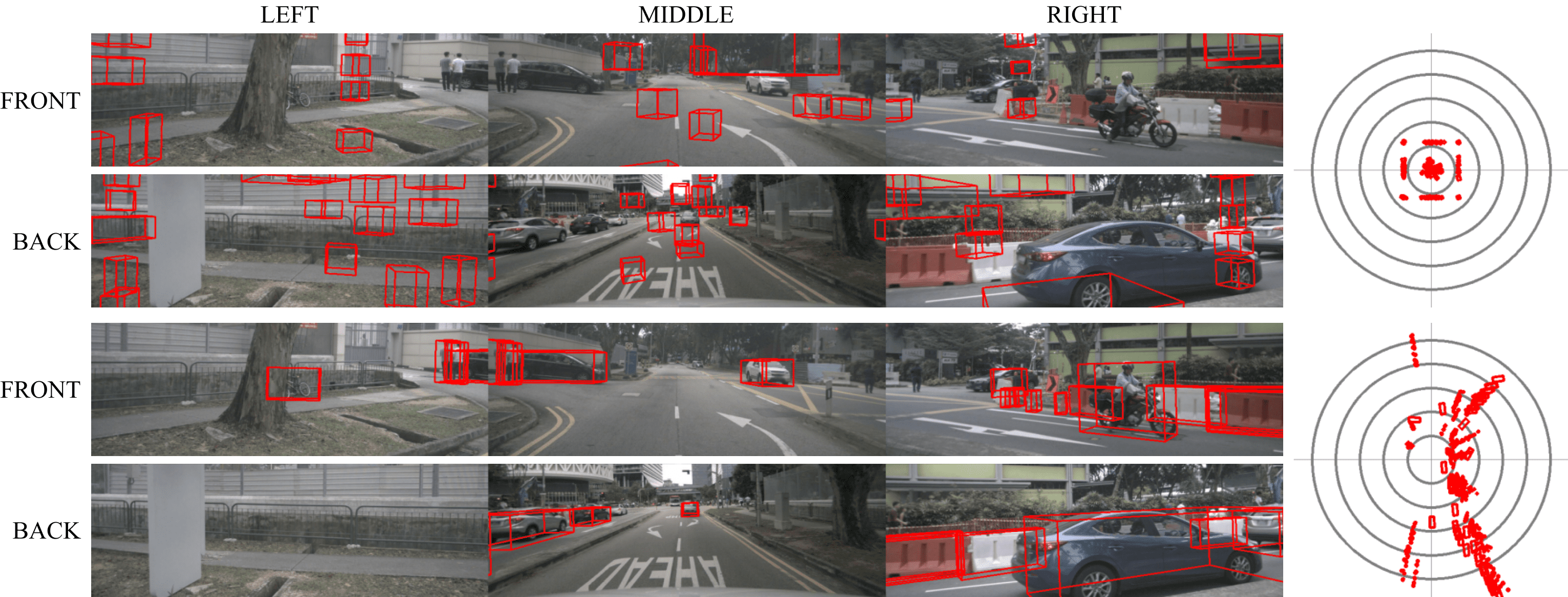}
    \caption{\textbf{Qualitative comparison between 3D proposals obtained from the learnable anchors setting (top) and our MDHA Anchor Encoder (bottom).} We visualize \textit{selected} proposals on all 6 cameras (left) and \textit{all} proposals in bird's-eye-view (right). For visual clarity, we display non-overlapping proposals for learnable anchors, and the top-20 proposals based on classification score for MDHA.}
    \label{fig:qual_learnable_vs_hybrid}
\end{figure*}
\subsubsection{Effectiveness of hybrid anchors}
\label{ssec: hybrid_vs_learnable_anchors}
To verify the effectiveness of our hybrid anchor scheme, we perform a quantitative and qualitative comparison with a learnable anchors baseline with no encoder, where proposals are implemented as parameterized embeddings with weights $W^{emb}$. Initial 3D anchor centers are obtained via $\sigma^{-1}(W^{emb})$ with zero-initialized queries. This setting mimics StreamPETR \cite{StreamPETR}.

Figure \ref{fig:qual_learnable_vs_hybrid} illustrates the comparison between proposals generated by these two approaches. In the learnable anchors setting, the same set of learned anchors are proposed regardless of the input scene, resulting in significant deviations from the actual objects. Thus, this approach relies heavily on the decoder to perform broad adjustments. It is also evident from the BEV that the proposals only cover a limited area around the vehicle, rendering them unsuitable for long-range detection. In contrast, our proposed method adapts to the input scene in two ways: first, the DepthNet predicts object depth based on image features; second, the Anchor Encoder selects only the top-$k$ proposals most likely to contain an object based on the feature map. Visually, our proposals not only manage to discriminate the relevant objects within the scene, they also encompass these objects quite well, detecting even the partially obstructed barriers in the back-right camera without explicit occlusion handling. Furthermore, as our proposals are unbounded, the BEV shows many proposals far away from the ego vehicle, making it more effective for detecting objects further away. These alleviate the decoder's load, allowing it to focus on fine-tuned adjustments and thus enhancing overall efficiency and performance.

These observations are consistent with the quantitative comparison in Table \ref{tab:quant_learnable_vs_hybrid}, where our hybrid scheme outperforms the learnable anchors baseline for \textit{all} metrics. Notably, it achieves a 7.1 pp improvement in mAP and a 5.8 pp improvement in NDS. Due to the limited range of the learnable anchors, it suffers from a large translation error, which was reduced by 12.4 pp in the hybrid anchors scheme.

\subsubsection{Ablation study on Circular Deformable Attention}
\label{ssec: ablation_cda}

Table \ref{tab:ablation_cda} shows that without view-spanning, multiple projected reference points (multi-projection) outperforms a single projected reference point (single-projection, Section \ref{ssec:decoder}) by 1.2 pp and 1.4 pp mAP and NDS. With view-spanning, although performance improves for both settings, multi-projection models obtain \textit{worse} mAP and NDS than their single-projection counterparts. The performance jump between models 4 and 5 validates the efficacy of our multi-view-spanning mechanism, and we hypothesize that the one-to-many relationship between queries and reference points in multi-projection models slows down convergence compared to a single well-chosen point, causing it to lag behind in performance. Moreover, multi-projection models are slower by 0.5 FPS due to the additional feature sampling. Wrapping sampling points around (wraparound) also yields a small performance increase. Both view-spanning and wraparound have no impact on the FPS. Thus, our CDA adopts the single-projection approach for improved efficiency, and  view-spanning with wraparound for maximal performance.

\begin{table}[t]
\begin{center}
\caption{Performance comparison when trained with proposals generated via learnable vs hybrid anchors\label{tab:quant_learnable_vs_hybrid}}
\begin{tabular}{P{0.15\textwidth}| P{0.034\textwidth} P{0.034\textwidth} P{0.0418\textwidth} P{0.0418\textwidth}  P{0.0418\textwidth}} 
\hline
\centering Method & mAP$\uparrow$ & NDS$\uparrow$ & mATE$\downarrow$ & mASE$\downarrow$ & mAVE$\downarrow$ \\
\hline
Learnable Anchors & 0.267 & 0.393 & 0.860 & 0.283 & 0.369 \\
Hybrid Anchors (Ours) & \textbf{0.338} & \textbf{0.451} & \textbf{0.736} & \textbf{0.268} & \textbf{0.365} \\
\rowcolor{BlueHighlight}
Improvement (pp) & 7.1 & 5.8 & 12.4 & 1.5 & 0.4 \\
\hline
\end{tabular}
\end{center}
\end{table}

\begin{table}[t]
\begin{center}
\caption{Ablation study on Circular Deformable Attention\label{tab:ablation_cda}}
\begin{tabular}{P{0.03\textwidth} | P{0.04\textwidth} P{0.04\textwidth} P{0.06\textwidth} P{0.036\textwidth} | P{0.035\textwidth} P{0.03\textwidth} | P{0.025\textwidth}} 
\hline
ID & Single-Proj. & Multi-Proj. & View-Spanning & Wrap & mAP$\uparrow$ & NDS$\uparrow$ & FPS$\uparrow$ \\
\hline
1 & & \checkmark & & & 0.313 & 0.416 & \multirow{3}{*}{13.8}\\
2 & & \checkmark & \checkmark & & 0.321 & 0.435 & \\
3 & & \checkmark & \checkmark & \checkmark & 0.323 & 0.438 & \\
\hline
\rule{0pt}{2.2ex}4 & \checkmark & & & & 0.301 & 0.402 & \multirow{3}{*}{\textbf{14.3}}\\
5 & \checkmark & & \checkmark & & 0.331 & 0.445 & \\
\rowcolor{BlueHighlight}\rule{0pt}{2.2ex}CDA &\checkmark & & \checkmark & \checkmark & \textbf{0.338} & \textbf{0.451} & \\
\hline
\end{tabular}
\end{center}
\end{table}

\begin{table}[h]
\begin{center}
\caption{Effect of number of sampling locations per reference point\label{tab:comp_sampling_locs}}
\begin{tabular}{ P{0.062\textwidth} P{0.062\textwidth} |P{0.034\textwidth} P{0.034\textwidth} P{0.0418\textwidth} P{0.0418\textwidth} P{0.0418\textwidth}} 
\hline
Encoder & Decoder & mAP$\uparrow$ & NDS$\uparrow$ & mATE$\downarrow$ & mAVE$\downarrow$ & FPS$\uparrow$\\
\hline
4 & 12 & 0.337 & 0.442 & 0.756 & \textbf{0.347} & \textbf{14.4} \\
12 & 12 & 0.328 & 0.446 & 0.740 & 0.357 & 13.8 \\
24 & 4 & 0.325 & 0.425 & 0.752 & 0.421 & 13.0 \\
\rowcolor{BlueHighlight}4 & 24 & \textbf{0.338} & \textbf{0.451} & \textbf{0.736} & 0.365 & 14.3 \\
\hline
\end{tabular}
\end{center}
\end{table}

\subsubsection{Number of sampling locations per reference point}
\label{ssec: num_sampling_locs}
Increasing the number of sampling locations per reference point in both the encoder ($S_{enc}$) and decoder ($S_{dec}$) enables queries to attend to more features. Table \ref{tab:comp_sampling_locs} shows that as $S_{enc}$ increases from 4 to 12, NDS improves by 0.4 pp, and as $S_{dec}$ increases from 12 to 24, both mAP and NDS improve by 0.1 pp and 0.9 pp, respectively. Comparing the results of $(S_{enc}, S_{dec})=(24, 4)$ with that of $(S_{enc}, S_{dec})=(4, 24)$, the latter achieves 1.3 pp and 2.6 pp higher mAP and NDS, while being faster by 1.3 FPS. Thus, increasing $S_{dec}$ yields a larger performance gain compared to increasing $S_{enc}$ by the same amount. This could be attributed to the difference in the number of queries, which is much higher in the encoder than the decoder. Hence, even with a low $S_{enc}$, the encoder's queries provide adequate input coverage. This also explains why increasing $S_{enc}$ results in more FPS reduction than increasing $S_{dec}$.
\section{CONCLUSION}
In this paper, we introduce MDHA, a novel framework which generates adaptive 3D output proposals using hybrid anchors. These proposals are sparsely refined and selected within our Anchor Encoder, followed by iterative refinement in the MDHA decoder. Sparse attention is performed by the multi-view-spanning CDA mechanism. MDHA significantly outperforms the learnable anchors baseline and achieves 46.4\% mAP and 55.0\% NDS on the nuScenes val set.

There are many possible improvements for MDHA. For instance, feature token sparsification \cite{focalpetr, sparse_detr} could enhance encoder efficiency, and the use of full 3D anchors \cite{sparse4d} as opposed to only 3D centers could improve performance. We hope that MDHA serves as a baseline for future advancements in query-based multi-camera 3D object detection.









\section*{ACKNOWLEDGMENT}
The work of Junn Yong Loo is supported by the Ministry of Higher Education Malaysia under the Fundamental Grant Scheme (FRGS) G-M010-MOH-000206.
\bibliographystyle{IEEEtran}
\bibliography{refs}

\end{document}